\documentclass[runningheads]{llncs}

\usepackage{graphicx}
\usepackage{amsmath,amssymb} 
\usepackage{color}
\usepackage[width=122mm,left=12mm,paperwidth=146mm,height=193mm,top=12mm,paperheight=217mm]{geometry}

\usepackage{cite}
\usepackage{array}
\usepackage{booktabs}
\usepackage{multirow}

\usepackage{fancyhdr}

\newcommand{\specialcell}[2][c]{%
	\begin{tabular}[#1]{@{}c@{}}#2\end{tabular}}

\begin{document}
\pagestyle{headings}
\mainmatter

\title{Towards Semantic Fast-Forward and Stabilized Egocentric Videos} 

\titlerunning{Towards Semantic Fast-Forward and Stabilized Egocentric Videos}

\authorrunning{Silva, M. \and Ramos, W. \and Ferreira, J. \and Campos, M. \and Nascimento, E. R.}

\author{Michel Melo Silva\thanks{First two authors contributed equally.} \and Washington Luis Souza Ramos\inst{\star} \and Joao Pedro Klock Ferreira \and Mario Fernando Montenegro Campos \and Erickson Rangel Nascimento}


\institute{Departamento de Ci\^encia da Computa\c{c}\~ao\\
	\email{\{michelms,washington.ramos,mario,erickson\}@dcc.ufmg.br}, \email{jpklock@ufmg.br}\\
	Universidade Federal de Minas Gerais, Brazil\\
}

\maketitle

\thispagestyle{fancy}
\fancyhf{}
\chead{\scriptsize{In Proceedings of the First International Workshop on Egocentric Perception, Interaction and Computing at European Conference on Computer Vision (EPIC@ECCV 2016) \\ The final publication is available at: http://dx.doi.org/10.1007/978-3-319-46604-0\_40}}
\setlength{\headsep}{0.35 in}

\begin{abstract}
The emergence of low-cost personal mobiles devices and wearable cameras and the increasing storage capacity of video-sharing websites have pushed forward a growing interest towards first-person videos. Since most of the recorded videos compose long-running streams with unedited content, they are tedious and unpleasant to watch. The fast-forward state-of-the-art methods are facing challenges of balancing the smoothness of the video and the emphasis in the relevant frames given a speed-up rate. In this work, we present a methodology capable of summarizing and stabilizing egocentric videos by extracting the semantic information from the frames. This paper also describes a dataset collection with several semantically labeled videos and introduces a new smoothness evaluation metric for egocentric videos that is used to test our method.
\end{abstract}

\section{Introduction}
\label{sec:introduction}

The popularity of first person videos, also known as egocentric videos, has considerably increased in social media. The large capacity of personal and video-sharing websites repositories and the ubiquity of easily operable devices such as smartphones, GoPro\texttrademark and Sony POV Action cameras are providing a compelling ecosystem for creating and storing different types of long-running egocentric videos. The wearer is free for recording long streams of regular activities such as working, cooking, driving, athletic activities like walking, running, bicycling, snowboarding, monitoring tasks (e.g. police patrol and life guarding) and home videos like family meetings and birthdays.

Despite the increasing popularity of recording egocentric videos, they are usually lengthy, monotonous and composed of an unedited content. The camera unsteadiness caused by the natural movements of the wearer makes them challenging to be analyzed \cite{pol_eph_pel_aro}. Sampling at a fixed rate is the most simple manner to reduce their length. However, it amplifies the body movements producing disturbing and even nauseates videos.

Several works have been proposed to address the instability of egocentric videos aiming to create a pleasant experience when watching the reduced version. Such works borrowed the term \textit{hyperlapse} from the exposure method in timelapse photography to name their methods. Similar to hyperlapse photography, where the camera moves through long distances and the images are aligned to create a final video with smooth transitions along the acquisition time, the hyperlapse algorithms also aim to downsize long and monotonous videos in short fast-forward watchable videos with no abrupt transitions between the frames. One challenge involving these approaches is that some portions of the video may be more significant to the users than others. For instance, one could be recording a celebratory event in its entirety, but in a posterior exhibition to the family the relevant parts are only those in which is possible to recognize the guests. 

In this work, we propose a new frame segmentation approach and an egocentric video stabilizer based on the disparity between the semantic and non-semantic segments. Our method minimizes the shakiness in the final video avoiding the deletion of relevant frames for the user based as far as the semantic information is concerned. A new dataset composed of semantically labeled videos and an evaluation metric to measure the egocentric videos smoothness are presented and used in  our experiments.

Similar to this work, our previous approach~\cite{ram_sil_cam_nas} slices the video into segments based on their relevance to the user to define their relative speed-ups. Although that approach is capable of creating a final video with the required speed-up avoiding the deletion of relevant segments, its optimization process increases the shakiness in the segments classified as no relevant. As stated by Poleg et al.~\cite{pol_hal_aro_pel} and Kopf et al.~\cite{kop_coh_sze}, egocentric videos do not present smooth transitions and continuous movement making hard to use traditional stabilization techniques which in the fast-forward videos is even more challenging due to the fact they are not composed of temporal consecutive frames. Additionally, like other works, the quantitative experiments is limited due to the use of a rough shakiness metric and an uncontrolled dataset, which generate misleading results.

\paragraph{Contributions:} The contributions of this work can be summarized as: 
\renewcommand{\theenumi}{\roman{enumi}}
\begin{enumerate}
	\item A new frame segmentation approach combined with an egocentric video stabilizer. Our method uses the disparity between the semantic and non-semantic parts to segment the input video and stabilizes the segments by using homography transformations to smooth the output video and reconstruct the frames;
	\item A new dataset with several semantically labeled videos to fill the gap in the literature related to well-controlled datasets concerning the semantic information;
	\item A new evaluation metric able to measure the egocentric videos smoothness. We demonstrate through qualitative results that the most used metric for this kind of video which is the reduction of epipole/Focus of Expansion (FOE) jitter is not accurate.
\end{enumerate}

\section{Related Work}
\label{sec:related_work}

Video Summarization methods can capture the essential information of the video and create a shorter version, thus the amount of time necessary to interpret the video content can be reduced \cite{yu_kan_mul, zhu_xia_wu}. The summarization methods are basically divided into two approaches: static storyboard or still-image abstract, where the most representative keyframes are selected to represent the video as a whole \cite{har_roo_wil},\cite{lee_gho_gra} and; dynamic video skimming or moving-image abstract, where a series of video clips compose the output \cite{ngo_ma_zha_2003},\cite{gon_liu}. Despite the large number of video summarization techniques proposed over the past years, only few works address summarization on egocentric videos \cite{lee_gho_gra},\cite{lu_gra},\cite{Potapov2014},\cite{Gygli2014}. Besides video summarization techniques aim to keep semantic information, it cannot give a temporal perception of the video, because some parts of the input video are completely removed \cite{pol_hal_aro_pel}.

We can roughly divide smooth fast-forward techniques into two categories: 3D model approaches, which consists of methods that, in short, reconstruct the scene structure and create an optimal path where a virtual camera would navigate and; 2D approaches, where methods basically work on selecting frames adaptively to compose the final video. The main advantage of the former category is the freedom with respect to the camera pose, however, on the other hand, they require camera parallax and the reconstruction step can take a while to be performed. The latter avoids the 3D reconstruction by skipping a subset of the input frames in order to maximize the smoothness of the output video. 

The work of Kopf et al.~\cite{kop_coh_sze} is an example of the 3D model category. The authors present a method which reconstructs the 3D input camera path by using structure from motion and per-frame proxy geometries and, performs an optimization in path location and orientation to create a virtual path and render the final video. Although remarkable results are presented, their technique requires significant camera motion and parallax and, in addition, demands a high computational cost.

More recent methods have adopted the optimization of frame selection~\cite{jos_kie_toe_uyt_coh},\cite{pol_hal_aro_pel},\cite{Halperin2016}. Poleg et al.~\cite{pol_hal_aro_pel} focus on an adaptive frame selection based on minimizing an energy function. They modeled the video as a graph by mapping the frames as the nodes and the edges weight reflecting the cost of the transition between the frames in the final video. The shortest path in the graph produces the best frames transitions for the final video composition. In the work of Joshi et al.~\cite{jos_kie_toe_uyt_coh} they present a more sophisticated algorithm which optimally selects frames from the input video as result of a joint optimization of camera motion smoothing and speed-up. They also perform a 2D video stabilization to create the hyperlapse result. Halperin et al.~\cite{Halperin2016} extended the work of Poleg et al. expanding the field of view of the output video 
by using a mosaicking approach on the input frames with single or multiple egocentric videos. 

Although the output videos of the aforementioned methods are appreciable, they are limited by the lack of considering the existence of scenes with different relevance for the recorder. In our previous work~\cite{ram_sil_cam_nas} we addressed this issue by slicing the video into semantic and non-semantic segments and, based on the length of the segments we control the playback speed of each type of segment. In order to decrease the shakiness still present in the output videos of~\cite{ram_sil_cam_nas}, which is caused by the increase of the playback speed in non-semantic segments, we propose in this work an egocentric video stabilizer which uses information from the original video. We also improve their slicing strategy to accurately define the semantic regions.

Despite the large number of proposed methods for video stabilization, they do not present good results for egocentric videos~\cite{kop_coh_sze_supplemental},\cite{pol_hal_aro_pel}. One example is the work of Hsu et al.~\cite{Hsu2012}. In their work the input video is segmented in patches with $\alpha$ length and then a single homography matrix is applied to all frames belonging to a given patch. The $\alpha$ value utilized was $2$ or $3$-seconds, which represents, for example, around a half of a minute in a $10\times$ fast-forward video. In this interval it is unlikely that all frames within a same patch are picturing the same scene, therefore it is impractical to find a homography consistency on them.

\section{Methodology}
\label{sec:methodology}

In the following two sections we detail our proposed methodology to create semantic smooth fast-forward videos.

\subsection{Semantic Egocentric Fast-Forwarding}
\label{subsec:methodology_frame_sampling}

\begin{figure}[t!]
	\centering
	\includegraphics[width=\textwidth]{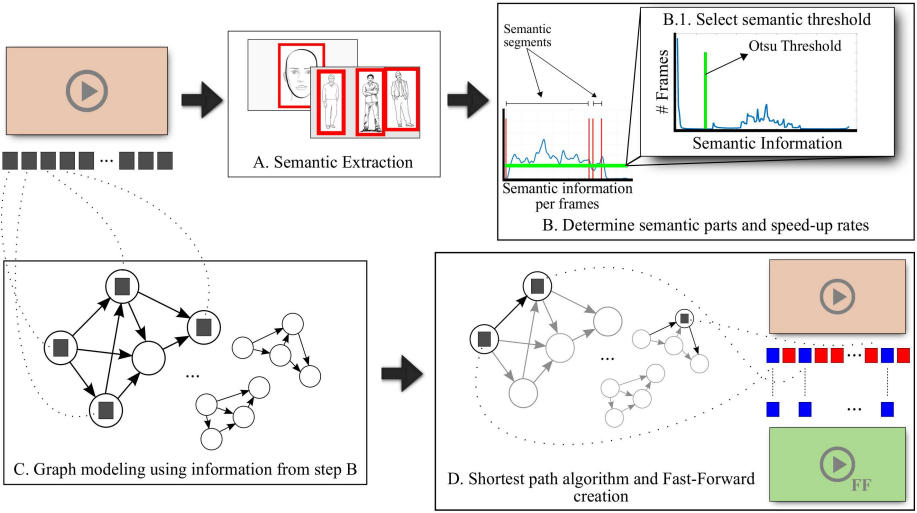}
	\caption{Frame sampling process with an egocentric video as input. A. Perform a semantic extraction to compute the relevance of the frames; B. Calculate the semantic score along the frames and B.1 uses the Otsu method to find a meaningful semantic threshold, in order to identify semantic parts and speed-up rates; C. Graphs are created from the frames and their relations; D. Compute the shortest path and compose the final video with the selected nodes.}
	\label{fig:methodology}
\end{figure}

The frame sampling process is composed of four steps. First, we build a graph for the egocentric video. In the graph frames are represented by nodes and the relation between two frames is modeled by an edge. Then, the semantic information, such as the Region of Interest (ROI) of detected faces or pedestrians, is extracted from each node in order to segment the video into relevant and non-relevant frames. Different speed-up rates are computed for relevant and non-relevant segments, such that the exhibition time of the semantic (the relevant segments) parts is enlarged to be contrasted over the non-semantic ones. At last, the final video is composed of the frames associated to the nodes in the shortest path in the graph. Figure~\ref{fig:methodology} summarizes our approach.

\paragraph*{Graph building:} Similar to the work of Poleg et al.~\cite{pol_hal_aro_pel}, we build the graph with each node connected with $\tau_{max}$ subsequent frames. The weight $W_{i,j}$ of the edge that connects the $i$-th to $j$-th node is given by the linear combination of the terms related to the frames transition instability, appearance, velocity and semantic multiplied by a proportional factor, as shown in Equation~\ref{eq:graph_formulation}:

\begin{equation}
	\label{eq:graph_formulation}
	W_{i,j} = (\lambda_{I} \cdot I_{i,j} + \lambda_{V} \cdot V_{i,j} + \lambda_{A} \cdot A_{i,j} + \lambda_{S} \cdot S_{i,j}) \cdot \left\lceil{\frac{(j-i)}{F}}\right\rceil,
\end{equation} 

\noindent where the proportional factor enhances transitions between frames with lower distance and $F$ is the speed-up rate applied in the graph which the edge belongs. 

The values of $\lambda$ coefficients are the regularization factors for each of the costs terms and $I_{i,j}$ is the Instability Cost Term, which is calculated as the average distance of the FOE to the center of the image. $V_{i,j}$ is the Velocity Cost Term which is given by the difference between the desired optical flow magnitude in the whole video and the average of the optical flow magnitudes sum along the consecutive frames from $i$ to $j$. $A_{i,j}$ is the Appearance Cost Term. We use a histogram comparison metric (Earth Mover's Distance) to measure the resemblance between the frame $i$ and $j$. 

\paragraph*{Semantic Extraction:} The Semantic Cost Term $S_{i,j}$ is used to penalize the transitions that are not composed by frames with relevant semantic information. Its values are given by Equation~\ref{eq:semantic_term}:  

\begin{equation}
	\label{eq:semantic_term}
	S_{i,j} = \frac{1}{S_{i} + S_{j} + \epsilon}.
\end{equation}

The value of $S_{x}$ presented in Equation~\ref{eq:semantic_term} is the semantic score of the frame $x$. For each frame of the input video, we extract the semantic information according to the semantic selected by the user (Fig.~\ref{fig:methodology}~A). Let $k$ be the $k$-th ROI returned by the extractor in the frame $x$, so the term $S_{x}$ is defined as follows:

\begin{equation}
	\label{eq:frame_semantic_score}
	S_{x} = \sum_{k \in f_{x}} c_{k} \cdot a_{k} \cdot G_{\sigma}(k),
\end{equation}

\noindent where $c_{k}$ is the classifier confidence about the ROI $k$, assigning relevance proportional to the reliability of the semantic information in frames; $a_{k}$ is the area of the $k$-th ROI, where ROIs with a bigger area represent a closer object, since it pictures a possible interaction and; $G_{\sigma}(k)$ is the value of the central point of the $k$-th ROI in the Gausian function with standard deviation $\sigma$ centered at the frame $f_x$, which returns higher values to centralized information, once it is an egocentric video, so the wearer is focused on the relevant information.

\paragraph*{Temporal Segmentation:} Using the semantic information along the frames, the video is segmented into semantic and non-semantic parts. Differently from our previous work~\cite{ram_sil_cam_nas}, that simply used the mean value (green line in Fig.~\ref{fig:methodology}~B) to determine the segmentation threshold, we apply in this work the Otsu thresholding method to find the threshold, running it in a histogram of semantic information. The value returned by Otsu (green line in Fig.~\ref{fig:methodology}~B.1) is used as the semantic threshold. Video segments composed of consecutive frames scored above this value are classified as semantic parts and the remaining ones as non-semantic.

\paragraph*{Speed-up rate estimation:} Estimating a lower speed-up rate for semantic segments and, consequently, a higher rate for non-semantic segments is not a trivial task, regarding their relation with the segment lengths. In order to manage the whole video in the desired speed-up $F_{d}$, the values of the semantic speed-up $F_{s}$ and the non-semantic $F_{ns}$ rates are computed by the minimization of the Equation~\ref{eq:DesiredSpeedUp}:

\begin{equation}
	\label{eq:DesiredSpeedUp}
	D(F_{ns}, F_{s}) = \left|\frac{L_{s}+L_{ns}}{F_{d}} - \left(\frac{L_{s}}{F_{s}} + \frac{L_{ns}}{F_{ns}} \right) \right|,
\end{equation}

\noindent where $L_{s}$ is the semantic segments length, in number of frames and, $L_{ns}$ is the non-semantic segments length.

We solve the Equation~\ref{eq:DesiredSpeedUp} by restricting the $F_{s}$ and $F_{ns}$ so that the $F_{s}$ value is minimized as well as the difference between both, as presented in Equation~\ref{eq:argmin}:

\begin{equation}
	\label{eq:argmin}
	\underset{F_{s},F_{ns}}{\arg\min} \left(D\left(F_{ns}, F_{s}\right) + \lambda_{1} \cdot |F_{ns}-F_{s}| + \lambda_{2} \cdot |F_{s}|\right),
\end{equation}

\noindent where $\lambda_{1}$ and $\lambda_{2}$ are the regularization terms that give more importance either to keep the speed-up rates close or take the smaller $F_{s}$.

We reduce the search space of the Equation~\ref{eq:argmin} by considering the restrictions: i) since we want more emphasis in the semantic parts, then $F_{s} \leqslant F_{d}$ and $F_{s} \leqslant F_{ns}$ and; ii) because $F_{s} \leqslant F_{d}$, in order to manage the final video speed-up rate, $F_{ns} \geqslant F_{d}$. Since $F_{ns}$, $F_{s}$ and $F_{d}\in \mathbb{N}$, then the search space is discrete and finite, due to the restrictions.

For each segment of the video we create a graph, with one source and one sink node, connecting with the $\tau_{b}$ border frames. For each graph we compute the shortest path through Bellman-Ford. All frames related with the nodes within the shortest path will compose the final video.

\subsection{Egocentric Video Stabilization}
\label{subsec:methodology_video_stabilization}

In this section we present a novel stabilization method for fast-forward egocentric videos, thus its input is the output of a frame sampling method. Instead of using Homography Consistency with smooth transitions like Hsu et al.~\cite{Hsu2012}, we propose to segment the video into patches and look for the master frame of each patch. We then create a transition area with the intermediate frames of every pair of masters. The key idea of our method is to create a smooth transition by setting the target image planes on the masters and modifying the image planes of the frames that belong to the transition areas.

\begin{figure}[t!]
	\centering
	\includegraphics[width=\textwidth]{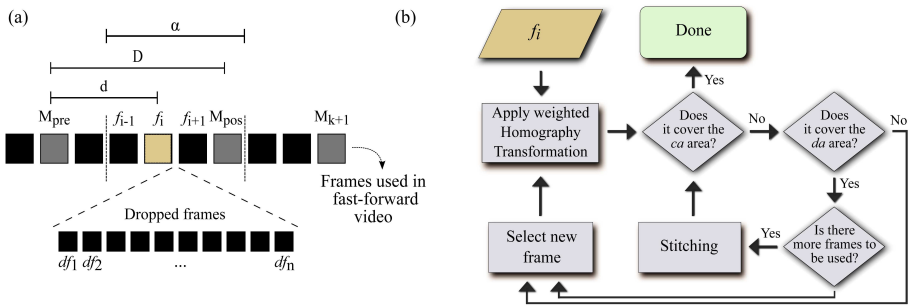}
	\caption{Stabilization methodology for fast-forward egocentric videos. (a)  Illustration of how the video is segmented in patches; dropped frames and the terms $\alpha$, $\Delta$ and $\delta$. (b) The diagram of the stabilization process.}
	\label{fig:stabilization_method}
\end{figure}

The first step of the stabilization methodology consists of segmenting the video into patches of size $\alpha$ and selecting one master frame $M_{k}$ for each patch (Fig. \ref{fig:stabilization_method}~a). We select as master of the $k$-th patch the frame $f$ that belongs to it and maximizes the Equation~\ref{eq:Find_Master_Frame}: 

\begin{equation}
	\label{eq:Find_Master_Frame}
	M_{k} = \underset{f}{\arg\max} \sum_{i \in p_{k}}R(f_i,f),
\end{equation}

\noindent where $p_{k}$ is the $k$-th patch and the $f_{i}$ is the $i$-th frame of the fast-forward video. The function $R(x,y)$ calculates the number of \textit{inliers} in the RANSAC method~\cite{Fischler1981} when computing the homography transformation from the image $x$ to $y$.

The second step is to smooth the transitions, similar to the work of Hsu et al. For each frame, we calculate two homography transformation matrices, one from the current frame to the previous master frame and another to the posterior one. Both homography transformations are applied with weights set according to the distance to the masters. The $i$-th frame of the stabilized video ($\widehat{f_{i}}$) is given by:

\begin{equation}
	\label{eq:smoothing_equation}
	\widehat{f_{i}} = H_{f_{i},M_{pre}}^{1-w} \cdot H_{f_{i},M_{pos}}^{w} \cdot f_{i},
\end{equation}

\noindent where $f_{i}$ is the $i$-th frame of the fast-forward video, $M_{pre}$ and $M_{pos}$ are respectively the previous and posterior master frames to the frame $f_{i}$; the term $H_{x,y}^{p}$ is the $p$-th power of the homography transformation matrix from the image $x$ to the image $y$; $w = (\delta \cdot (2 \cdot \alpha) / \Delta)$ is the weight that composes the $p$-th power, where $\delta$ is the distance (in number of frames) from the frame $f_{i}$ to $M_{pre}$, and $\Delta$ is the distance between $M_{pre}$ and $M_{pos}$ (Fig.~\ref{fig:stabilization_method}~a). As stated by Hsu et al., choosing the $\alpha$ value to be a power of 2 makes the root calculation feasible by consecutive square roots.

As expected, after applying the homography transformations estimated in Equation~\ref{eq:smoothing_equation}, black areas are generated due to the fact that the camera movements are abrupt and the elapsed time between consecutive frames in the fast-forward egocentric videos are large. Thus, the last step is to reconstruct these corrupted regions. To reconstruct these frames, we define two image areas centered in the frame: i) the drop area ($da$) equals to $dp\%$ size of the frame and; ii) the crop area ($ca$) equals to $cp\%$ size of the frame, where $cp > dp$.

The $da$ area is the center of the image, where the viewer focuses on the majority of the time, then it is not allowed have any black or reconstructed areas in this region. On the other hand, the area between the $ca$ and $da$ is the peripheral vision, which is allowed to have artifacts but not black areas. The $ca$ area is the cut region, thus regions outside this area are removed in the final video, therefore, having these black areas within them does not cause issues.

The reconstruction procedure is an iterative process depicted in Fig.~\ref{fig:stabilization_method}~(b). It starts with a new image $f_{i}$ as input. Firstly, the algorithm applies the weighted homography transformation (Eq.~\ref{eq:smoothing_equation}) resulting the $\widehat{f}_{i}$. Then, it checks if the $\widehat{f}_{i}$ covers the $ca$ area. If it does, no further actions are required and this frame is ready to compose the new stabilized video; otherwise the algorithm verifies if the $\widehat{f}_{i}$ covers the $da$ area. If it does not, the frame $\widehat{f}_{i}$ is dropped and a new image is selected, if it covers, the algorithm checks weather still exist unused frames skipped by the frame sampling process to perform the reconstruction process. In an affirmative case, one new skipped frame is selected and used in the stitching process. Otherwise, the $\widehat{f}_{i}$ is dropped out and a new frame needs to be selected. Whenever a new frame is selected the process starts again.

The stitching step is performed as follows. We use the SURF detector to select feature points in the frames $\widehat{f_{i}}$ and $d_{j}$. To calculate the homography transformation matrix we match feature points between the images by describing all feature points of $d_{j}$ and $\widehat{f_{i}}$ with SIFT descriptors and applying the brute force matching strategy. With the matched points we calculate the homography matrix $H_{d_{j},\widehat{f_{i}}}$ using the RANSAC method. The $H_{d_{j},\widehat{f_{i}}} \cdot d_{j}$ is now aligned with $\widehat{f}_{i}$ and copied to the back of it to compose the stitched image.

If it is necessary to select a new frame, it means that the $\widehat{f}_{i}$ does not yield a good transition in the final video. The algorithm selects a new frame $d_{j}$ that belongs to interval $[f_{i-1}, f_{i+1}]$ in the original video and maximizes the Equation~\ref{eq:select_new_frame}:
\begin{equation}
	\label{eq:select_new_frame}
	\underset{d_{j}}{\arg\max}~(~G_{\sigma}(p) \cdot (~R(d_{j},\widehat{f}_{i-1}) + R(d_{j},\widehat{f}_{i+1})~) \cdot (\eta + S(d_{j}))~),
\end{equation}
\noindent where, $G_{\sigma}(x)$ is the value of the Gaussian function with mean zero and standard deviation $\sigma$ in the position $x$; $p$ is the $ar$ area percentage covered by $d_j$; $\eta$ is a value used to prevent multiplication by zero, in case the function $S(d_{j})$ that calculate the semantic information in the frame $d_{j}$ returns zero. The final stabilized video is composed by all frames that achieved the Done step.

\section{Experiments}
\label{sec:experiments}

In this section we present the experimental evaluation and results for our methodology using the new dataset and the evaluation metric. The next two sections contain details about our contributions: the dataset composition and the shakiness metric.

\subsection{Semantic Egocentric Dataset}
\label{subsec:experiments_semantic_egocentric_dataset}

We propose a new labeled dataset to run the experiments and validate our methodology since there are no datasets in the literature that are semantically controlled.
The dataset is composed of $11$ videos divided in $3$ categories of different activities: Biking; Driving and Walking. The videos under each one of these categories are classified according to their amount of semantic information. The classes are: 0p, which represents the videos with approximately no semantic information present (Biking 0p, Driving 0p and Walking 0p); 25p, for the videos containing relevant semantic information in $\sim$25\% of its frames (Biking 25p, Driving 25p and Walking 25p); 50p, for the ones with around a half of their frames composed by semantics (Biking 50p, Biking 50p2, Driving 50p and Walking 50p) and; 75p, which represents videos with $\sim$75\% of their frames containing relevant semantic information (Walking 75p).

We selected sections where the semantic was present to record the videos. We computed the semantic information in frames, according to Equation~\ref{eq:frame_semantic_score}, by either using the NPD (Normalized Pixel Difference) Face Detector~\cite{lia_jai_li} for the videos of the Walking category or, a pedestrian detector~\cite{pdollarMATLAB} for the videos of the other categories. We intended to use faces as the semantic information for all videos, but the usage of the pedestrian detector was necessary since the videos when biking or driving present a higher motion speed, what prevent the face detector from achieving a substantial accuracy.

\begin{figure}[t!]
	\centering
	\includegraphics[width=\textwidth]{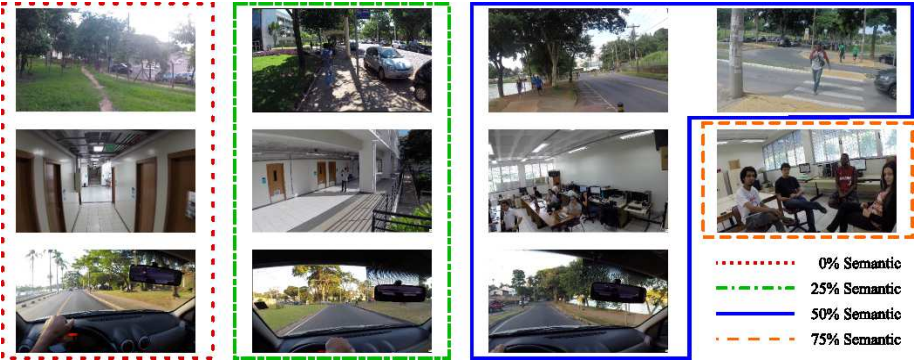}
	\caption{Examples of the proposed semantic egocentric dataset. Frames in the first row represent the videos of the Biking category. Frames in the second row represent the videos of the Walking category. Frames in the third row represent the videos of the Driving category.}
	\label{fig:dataset}
\end{figure}

We used a GoPro\texttrademark Hero 3 camera mounted in a helmet for the Biking and Walking videos and attached to a head strap for the Driving videos. All videos were recorded in daylight so that the detectors could achieve a better accuracy. Fig.~\ref{fig:dataset} shows some frame examples of the sequences in the dataset. The complete dataset, including videos and the semantic labels, is publicly available to the research community~\footnote{www.verlab.dcc.ufmg.br/fast-forward-video-based-on-semantic-extraction}.

\subsection{Shakiness Evaluation Metric}
\label{subsec:Shakiness_Metric}

Most of hyperlapse methodologies focuses on producing smooth fast-forward egocentric videos. In order to evaluate the smoothness of the output videos we need an evaluation metric that accurately express this value. The most popular quantitative measure present in the literature is the reduction of the epipole/FOE jitter~\cite{pol_hal_aro_pel},\cite{Halperin2016},\cite{ram_sil_cam_nas}. However, this metric assigns a higher score for some videos that are visually more shaky than others. Based on that, we conducted an user study to verify the real smoothness of the video and to assess the quality of the metric.

Inspired by the qualitative comparison between videos made by Joshi et al.~\cite{jos_kie_toe_uyt_coh}, where they made side-by-side comparisons using only the mean and standard deviation of consecutive output frames~\cite{jos_kie_toe_uyt_coh}, we devised a quantitative metric to calculate the smoothness of the video. We use the fact that the presence of sharper images indicates a more stable video.

Thus, the smoothness estimation is computed as:

\begin{equation}
	\label{eq:instability_index}
	I = \frac{1}{N} \cdot \sum_{i=1}^{N}~\frac{\sum_{j \in B_i}~(f_{j} - \bar{f_{i}})^2}{(N_{B}-1)},
\end{equation}

\noindent where $N$ is the number of frames of the video, $B_i$ is the $i$-th buffer composed by $N_{B}$ temporal neighborhood frames, $f_j$ is the $j$-$th$ frame of the video, $\bar{f_i}$ is the average frame of the buffer $B_i$ and $I$ indicates the instability index of the video. A smoother video yields a smaller $I$ value.

\begin{figure}[t!]
	\centering
	\includegraphics[width=0.75\textwidth]{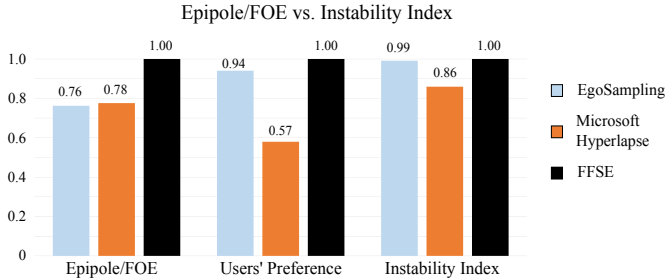}
	\caption{Comparison among the epipole/FOE metric, the users' preference and the instability metric. The epipole/FOE metric present a low mean for the ES algorithm, which does not match the users' preference, differently from our proposed metric which seems to be a better match.}
	\label{fig:foe_vs_instability}
\end{figure}

For the qualitative evaluation, we generated output videos with average length of $35$ seconds from $9$ sequences using the smooth fast-forwarding techniques: EgoSampling (ES)~\cite{pol_hal_aro_pel}; Microsoft Hyperlapse (MH)~\cite{jos_kie_toe_uyt_coh} and Fast-Forward Based on Semantic Extraction (FFSE)~\cite{ram_sil_cam_nas}, with a speed-up factor of $10$. These sequences are publicly available and were previously used by those works. Then, we asked for $33$ subjects to watch the (unlabeled) videos and grade the video instability with respect to its smoothness in an assessment questionnaire. Unlike the quantitative measure of FOE locations differentiation, where the ES technique is superior to the other two techniques, the majority of the subjects preferred watching the MH output video, as shown in Figure~\ref{fig:foe_vs_instability}. 

Figure~\ref{fig:foe_vs_instability} shows the normalized mean values of the $9$ sequences for the metrics. The results reveal that the proposed metric really reflects the subjects' preferences, since it is more similar.

\subsection{Results}

The results of our methodology are presented in this section. We conducted our experiments using the whole semantic egocentric dataset proposed in Section~\ref{subsec:experiments_semantic_egocentric_dataset}.

In our work, we used as the semantic extractors a face detector~\cite{lia_jai_li} in videos where the wearer is walking and a pedestrian detector~\cite{pdollarMATLAB} in videos where the motion speed is higher. These detectors are responsible for giving us the $c_k$ value of the Equation~\ref{eq:frame_semantic_score}. The Gaussian function is a Normal with parameters $\mu=0$ and $\sigma=min(W/2, H/2)$, where $W$ is the frame width and $H$ is the frame height. The maximum allowed skip was set to $\tau_{max} = 100$ and $\epsilon = 1$ is the value which prevents division by zero in Equation~\ref{eq:semantic_term}. The $\lambda$ values in Equations~\ref{eq:graph_formulation} and~\ref{eq:argmin} and the value of $\eta$ in Equation~\ref{eq:select_new_frame} were empirically defined, as well as the drop area, set as dp=50\% of the frame and the crop area, set as cp=5\% of the frame.

Our results were compared to three different approaches: i) the EgoSampling (ES) technique proposed by~\cite{pol_hal_aro_pel}. We used their best reported parameters; ii) the Microsoft Hyperlapse (MH) proposed by~\cite{jos_kie_toe_uyt_coh}, which we used the desktop version of their algorithm to generate the videos and; iii) the technique proposed in our previous work~\cite{ram_sil_cam_nas} (FFSE). All comparisons were made under quantitative metrics with respect to the visual instability, length and semantic information present in the output video. The metrics are:
\begin{enumerate}
	\item[1.] Output Speed-up: this metric indicates the speed-up achieved by the output video. It is better to have a speed-up close to the required speed-up. In this work, we set the speed-up to the factor of 10.
	\item[2.] Semantic Content: it is the sum of the $S_x$ value for all frames $f_x$ of the video. We want to get a higher value over the other techniques.
	\item[3.] Instability Index: this is the value of the Equation~\ref{eq:instability_index}. The lower is this value, the smoother is the video.
\end{enumerate}

\begin{figure}[t!]
	\centering
	\includegraphics[width=0.75\textwidth]{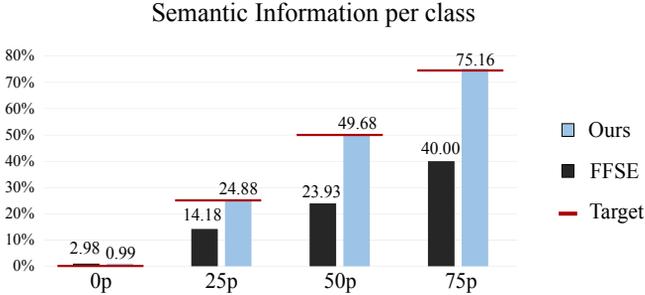}
	\caption{The mean percentage of the semantic information present in videos per class. The values show how close the algorithms are from the ideal (target) value for the class. A peak in a frame with semantic information makes the FFSE segmentation strategy achieve poor results, differently from ours.}
	\label{fig:semantic_segmentation}
\end{figure}

Figure~\ref{fig:semantic_segmentation} shows the mean percentage of semantic information present in videos per class for the FFSE algorithm and ours. We expect the methods to have the value as close as possible to the class. For instance, the methods should yield 25\% of semantics for the class 25p. The values obtained by our algorithm are closer to the expected values, which means that our segmentation strategy is more accurate. The reason of the poor accuracy of the FFSE algorithm is that a peak in semantic information interferes directly their approach, since they use the mean. In our case it does not happen, because we use a method that analyzes the semantic information globally to find the better threshold value.

\begin{table}[h!]
	\centering
	\caption{Semantic Content}
	\begin{tabular}{>{\arraybackslash}m{2.2cm}>{\raggedleft\arraybackslash}m{1.6cm}>{\raggedleft\arraybackslash}m{1.6cm}>{\raggedleft\arraybackslash}m{1.6cm}>{\raggedleft\arraybackslash}m{2cm}}
		\toprule
		\textbf{Name} & \textbf{ES} & \textbf{MH} & \textbf{FFSE} & \textbf{Ours} \\
		\midrule
		Biking 0p     & \textbf{142.15}        & 54.47                 & 114.76   & 114.76     \\
		Biking 25p    & 1,832.47       & 3,517.28               & 3,527.81  & \textbf{3,758.44}    \\
		Biking 50p    & 4,640.36       & 3,374.42               & 6,247.30  & \textbf{6,713.49}    \\
		Biking 50p2 & 2,650.41       & 4,760.14               & \textbf{6,955.12}  & 5,744.19    \\
		Driving 0p    & 13.12         & 24.66                 & \textbf{39.70}    & \textbf{39.70}      \\
		Driving 25p   & 121.76        & 216.17                & 220.69   & \textbf{238.18}     \\
		Driving 50p   & 228.38        & 533.95                & 479.35   & \textbf{569.66}     \\
		Walking 0p    & 161.87        & 239.28                & \textbf{259.69}   & \textbf{259.69}     \\
		Walking 25p   & 6,655.91       & 30,553.46              & \textbf{78,752.01} & 69,703.28   \\
		Walking 50p   & 5,817.10       & 3,497.56               & 51,603.23 & \textbf{53,700.62}   \\
		Walking 75p   & 16,594.22      & 72,883.23              & 93,074.20 & \textbf{125,766.91} \\
		\bottomrule
	\end{tabular}%
	\label{tab:semantic_content}%
\end{table}%

We calculated the speed-up rates for the output videos. The ES algorithm reported a mean value of 23.904 and a standard deviation of 6.91, which is far from the ideal. The FFSE and our proposed technique achieved mean values of 12.764 and 12.188 and, standard deviation of 2.53 and 2.83 respectively, which means that they have a better accuracy with respect to the ES algorithm. The MH is the better one, because it achieves a mean of 9.44, which is the closest value to the required speed-up rate, and a standard deviation of 1.41.

Table~\ref{tab:semantic_content} presents the results for the semantic content metric. Our algorithm is better than the others in the most of the cases. It is worth to note that for the videos with low semantic information present, our technique manages keep the same semantic information reported by the FFSE algorithm, which enforces our improvement with respect to the semantic.

We show in Table~\ref{tab:instability_index} results of the instability index metric for all approaches where lower values denote better results. We added a column to report the results for our stabilized videos in which we can see the improvement in the smoothness of the output videos.

\begin{table}[h!]
	\centering
	\caption{Instability Comparison}
	\begin{tabular}{>{\arraybackslash}m{1.8cm}>{\raggedleft\arraybackslash}m{1.6cm}>{\raggedleft\arraybackslash}m{1.6cm}>{\raggedleft\arraybackslash}m{1.6cm}>{\raggedleft\arraybackslash}m{1.6cm}>{\raggedleft\arraybackslash}m{2.3cm}}
		\toprule
		\textbf{Name} & \textbf{ES} & \textbf{MH} & \textbf{FFSE} & \textbf{Ours} & \textbf{\specialcell{Ours +\\Stabilization}}\\
		\midrule
		Biking 0p     & 111.70        & 84.70                 & 113.53 & 113.53   & \textbf{81.76}                \\
		Biking 25p    & 187.25        & 166.77                & 185.57 & 185.03   & \textbf{128.02}               \\
		Biking 50p    & 134.22        & 112.95                & 132.95 & 129.97   & \textbf{90.68}                \\
		Biking 50p2 & 107.59        & 95.82                 & 110.64 & 111.71  & \textbf{76.75}                \\
		Driving 0p    & 164.51        & 153.33                & 177.60 & 177.60   & \textbf{127.02}               \\
		Driving 25p   & 148.25        & 137.75                & 152.42 & 152.18   & \textbf{124.07}               \\
		Driving 50p   & 154.38        & 131.06                & 154.82 & 153.85   & \textbf{109.29}               \\
		Walking 0p    & 126.33        & 121.21                & 133.20 & 133.20   & \textbf{94.26}                \\
		Walking 25p   & 134.96        & 126.06                & 129.73 & 132.49   & \textbf{96.23}               \\
		Walking 50p   & 139.45        & 119.06                & 138.70  & 138.62   & \textbf{92.53}                \\
		Walking 75p   & 150.18        & 127.55                & 145.75 & 137.15   & \textbf{99.78}                \\
		\bottomrule
	\end{tabular}%
	\label{tab:instability_index}%
\end{table}%

\section{Conclusions}
\label{sec:conclusions_and_future_works}

In this work we proposed a novel method capable of producing smoother egocentric videos with more semantic content, by considering the shakiness, the required speed-up and the semantic information. We also introduced a new semantically controlled dataset and a smoothness evaluation metric to test fast-forward egocentric methods, once most of metrics in the literature do not reflect the watchers' preferences. We ran several experiments using the new dataset and the evaluation metric. The results showed the superiority of our new approach as far as smoothness and semantic information are concerned.

\paragraph{\textbf{Acknowledgments.}}
The authors would like to thank the agencies CAPES, CNPq, FAPEMIG, ITV (Vale Institute of Technology) and Petrobras for funding different parts of this work. 

\bibliographystyle{splncs03}
\bibliography{EPIC_ECCV_2016_cr}
\end{document}